\documentclass[letterpaper, 10 pt, journal, twoside]{IEEEtran}

\usepackage{amsmath,amsfonts}
\usepackage{algorithmic}
\usepackage{array}
\usepackage[caption=false,font=normalsize,labelfont=sf,textfont=sf]{subfig}
\usepackage{textcomp}
\usepackage{stfloats}
\usepackage{url}
\usepackage{verbatim}
\usepackage{graphicx}
\def\BibTeX{{\rm B\kern-.05em{\sc i\kern-.025em b}\kern-.08em
    T\kern-.1667em\lower.7ex\hbox{E}\kern-.125emX}}
\usepackage{balance}

\usepackage{graphics} 
\usepackage{epsfig} 
\usepackage{mathptmx} 
\usepackage{times} 
\usepackage{amsmath} 
\usepackage{amssymb}  
\usepackage{physics}
\usepackage{amsmath}
\usepackage{cite}
\usepackage{cleveref}
\usepackage{tabularx}
\usepackage{multirow}
\usepackage{algorithm}
\usepackage{algorithmic}
\usepackage{booktabs}
\usepackage{xcolor}
\usepackage{makecell}
\DeclareMathAlphabet\mathzapf       {T1}{pzc} {mb} {it}
\usepackage{array}

\DeclareGraphicsExtensions{.pdf,.jpeg,.png}

\Crefname{figure}{Fig.}{Figs.}

\begin{document}
\title{Hysteresis Compensation of Flexible Continuum Manipulator using RGBD Sensing and Temporal Convolutional Network}

\author{Junhyun Park$^{1*}$, Seonghyeok Jang$^{1*}$, Hyojae Park$^{1}$, Seongjun Bae$^{1}$, and Minho Hwang$^{1\dagger}$
\thanks{Manuscript received: January 8, 2024; Revised: April 7, 2024; Accepted: April 30, 2024.} 
\thanks{This paper was recommended for publication by Editor Jens Kober upon evaluation of the Associate Editor and Reviewers’ comments.}
\thanks{This work was supported in part by the DGIST R\&D Program of the Ministry of Science and ICT (23-PCOE-02, 23-DPIC-20) and by the collaborative project with ROEN Surgical Inc. This work was supported in part by the Korea Medical Device Development Fund grant funded by the Korea government (the Ministry of Science and ICT, the Ministry of Trade, Industry and Energy, the Ministry of Health \& Welfare, the Ministry of Food and Drug Safety) (Project Number: 1711196477, RS-2023-00252244) and by the National Research Council of Science \& Technology (NST) grant by the Korea government (MSIT) (CRC23021-000).}
\thanks{* These authors contributed equally, † Corresponding author.} 
\thanks{$^1$ Junhyun Park, Seonghyeok Jang, Hyojae Park, Seongjun Bae, and Minho Hwang are with the Department of Robotics and Mechatronics Engineering, DGIST, Daegu, 42988 Korea (e-mail: {\tt\footnotesize \{sean05071, jshtop1, hyojae, twin010528, minho\} @dgist.ac.kr}).}%
\thanks{Digital Object Identifier (DOI): see top of this page}
}

\markboth{IEEE ROBOTICS AND AUTOMATION LETTERS. PREPRINT VERSION. ACCEPTED APRIL, 2024}%
{Park \MakeLowercase{\textit{et al.}}: Hysteresis Compensation of Flexible Continuum Manipulator} 

\maketitle

\begin{abstract}
Flexible continuum manipulators are valued for minimally invasive surgery, offering access to confined spaces through nonlinear paths. However, cable-driven manipulators face control difficulties due to hysteresis from cabling effects such as friction, elongation, and coupling. These effects are difficult to model due to nonlinearity and the difficulties become even more evident when dealing with long and coupled, multi-segmented manipulator. This paper proposes a data-driven approach based on Deep Neural Networks (DNN) to capture these nonlinear and previous states-dependent characteristics of cable actuation. We collect physical joint configurations according to command joint configurations using RGBD sensing and 7 fiducial markers to model the hysteresis of the proposed manipulator. Result on a study comparing the estimation performance of four DNN models show that the Temporal Convolution Network (TCN) demonstrates the highest predictive capability. Leveraging trained TCNs, we build a control algorithm to compensate for hysteresis. Tracking tests in task space using unseen trajectories show that the proposed control algorithm reduces the average position and orientation error by 61.39\% (from $\mathbf{13.7mm}$ to $\mathbf{5.29 mm}$) and 64.04\% (from 31.17$^{\circ}$ to 11.21$^{\circ}$), respectively. This result implies that the proposed calibrated controller effectively reaches the desired configurations by estimating the hysteresis of the manipulator. Applying this method in real surgical scenarios has the potential to enhance control precision and improve surgical performance.
\end{abstract}

\begin{IEEEkeywords}
Tendon/Wire Mechanism, Machine Learning for Robot Control, Modeling, Control, and Learning for Soft Robots
\end{IEEEkeywords}

\section{Introduction}

\IEEEPARstart{I}{n} contrast to rigid surgical robots that navigate lesions through laparoscopic approaches, flexible endoscopic surgical robots can access endoluminal regions by traversing a curved path through natural orifices such as the mouth, anus, and vagina. Due to their heightened accessibility to lesions and clinical advantages of scar-free procedures, a new robotic platform has been actively under investigation \cite{Hwang2020KFLEXAF, Remacle2015TransoralRS, Phee2009MasterAS, Zorn2018ANT}. These platforms have demonstrated effectiveness in endoscopic tasks like endoscopic submucosal dissection. However, they lack the necessary degrees-of-freedom (DOF) to execute motions for complex tasks such as tissue suturing and vascular anastomosis. Despite the need for additional DOFs to perform more advanced tasks, challenges arise in designing multi-DOF due to size constraints \cite{daVeiga2020ChallengesOC}.

Another significant limitation of the existing platforms is in control inaccuracy due to long and flexible actuation cables. Endoscopic surgical platforms are remotely powered by multiple bundles of Bowden-cables passing through a lengthy flexible tube, exceeding 1.5m in length. The prolonged cable introduces challenges such as friction, twisting \cite{Ji2020AnalysisOT}, extension \cite{Dalvand2018AnAL}, backlash \cite{Kim2020EffectOB}, and coupling \cite{Roy2017ModelingAE}, leading to uncertainties in the control and hysteresis of the flexible manipulator. 
\begin{figure}[t!]
    \centering
    \includegraphics[width=0.9\linewidth]{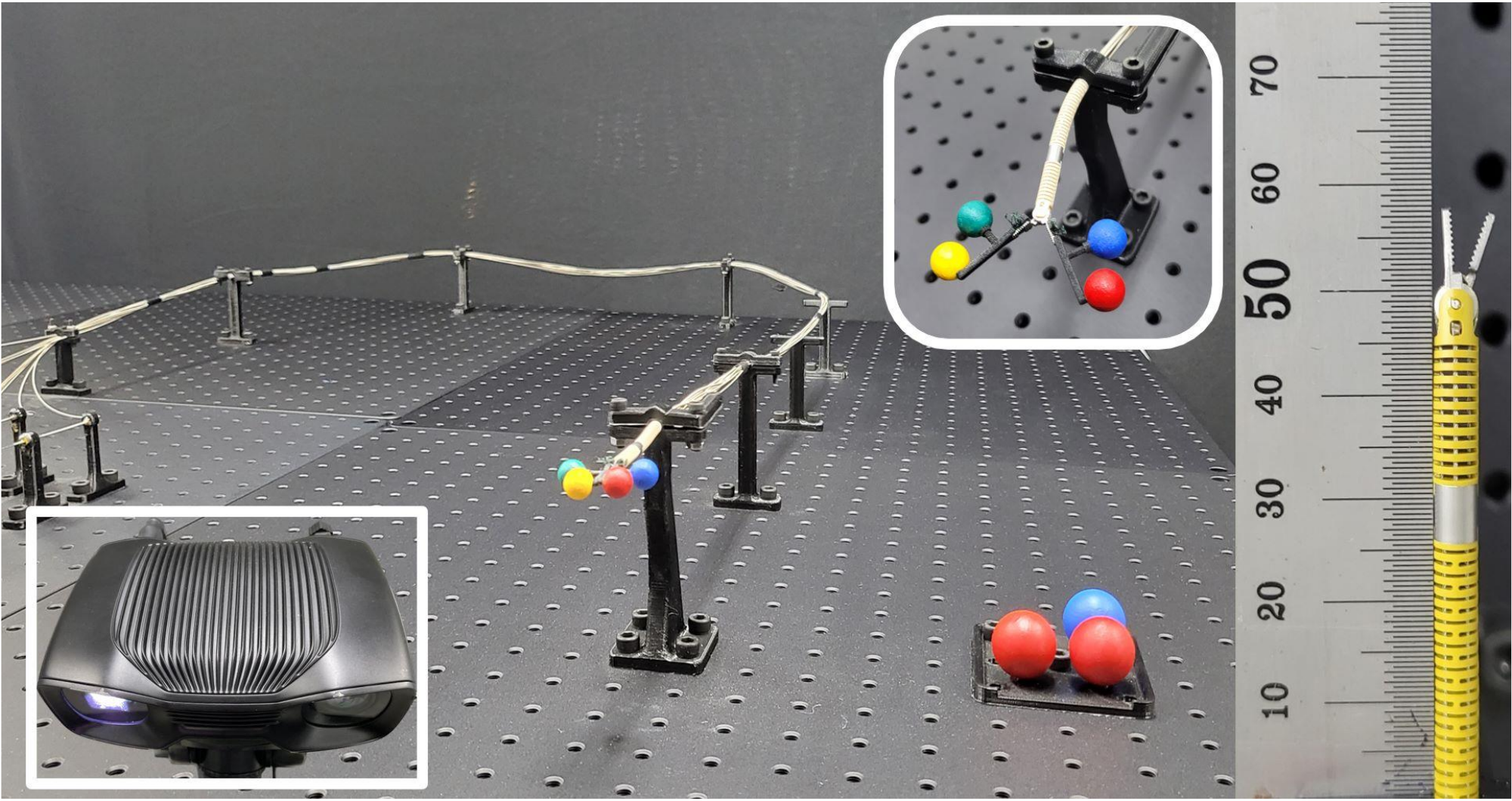}
    \vspace{-1em}
    \caption{The 3D printed fiducial markers are attached to the forceps to estimate the physical joint angles of the proposed continuum manipulator. We use a RGBD camera to detect the central position of fiducial marker.}
    \label{demo_figure}
    \vspace{-0.5em}
\end{figure}

This letter focuses on overcoming the challenges associated with flexible manipulators. To address the constraints imposed by insufficient DOFs, we propose the design of a dual-segment continuum effectively perform endoscopic surgery within confined anatomical spaces. This study includes kinematic analysis and derivation of the cable drive equation associated with the proposed manipulator.

To improve control accuracy, we adopt a data-driven approach. 7 fiducial markers are attached to capture physical poses of the proposed manipulator using RGBD sensing (See Fig. \ref{demo_figure}), collecting data to identify hysteresis. The dataset is used to train deep learning models with 2 different approaches, 4 distinct architectures, and varying lengths of input sequences. Utilizing the TCN models with sequence lengths of 80, we design a control algorithm to compensate for hysteresis, returning the calibrated command joint angle to reach the inputted desired joint angle.

The main contributions of this paper are summarized as follows: (1) Design of a dual-segment manipulator for flexible endoscopic surgery. (2) Estimation of the physical joint configuration of the manipulator, including forceps, based on fiducial markers and RGBD sensing. (3) Proposal of a hysteresis compensation algorithm employing TCN models to calibrate hysteresis effects in the proposed long and flexible continuum manipulator. (4) Validation through 3 unseen trajectories tracking test suggesting that the proposed control algorithm can significantly reduce the hysteresis effects.

\begin{figure}[t!]
    \centering
    \includegraphics[width=0.8\linewidth]{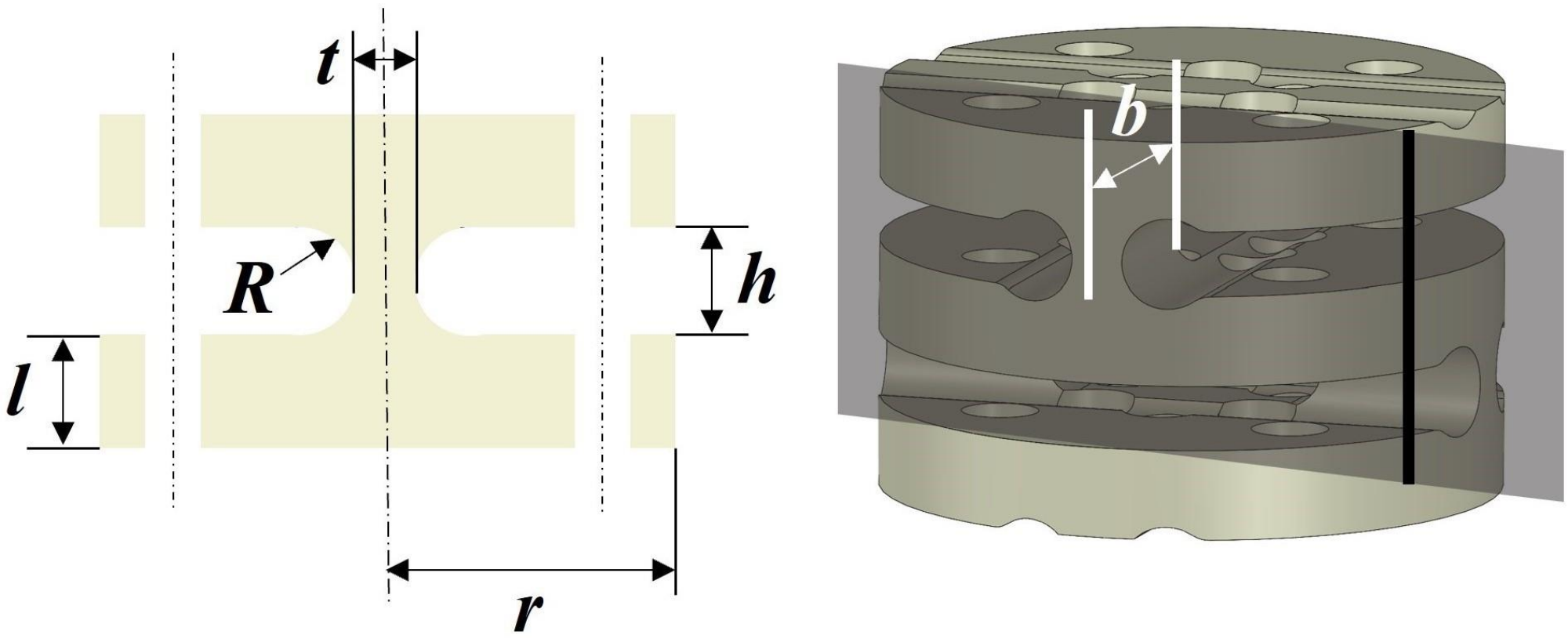}
    \vspace{-1.1em}
    \caption{\textbf{Design Parameters of Flexure Hinge Module:} The design of the flexure hinges in the manipulator is based on the Circular Flexure Hinge Design by Paros and Weisbord \cite{Tseytlin2002NotchFH}.}
    \vspace{-0.7em}
    \label{Design_para}
\end{figure}

\begin{table}[t!]
\caption{Design Parameters of Flexure Hinge Module}
\vspace{-1.4em}
\label{tab:Design_para}
\begin{center}
\setlength{\tabcolsep}{45pt}
\renewcommand{\arraystretch}{1.3}
\begin{tabular}{@{\hspace{10pt}} ll @{\hspace{10pt}}}
\toprule
\textbf{Parameters}            & \textbf{Value}  \\ 
\midrule
Height between notches, \textit{h}      & $\mathrm{0.5mm}$  \\
Thickness of the disk, \textit{l}       & $\mathrm{0.7mm}$  \\
Round radius of the hinge, \textit{R}   & $\mathrm{0.35mm}$ \\
Front thickness, \textit{t}             & $\mathrm{0.4mm}$  \\
Side thickness of the hinge, \textit{b} & $\mathrm{1.3mm}$  \\
Radius of module, \textit{r}            & $\mathrm{2.4mm}$  \\ 
\bottomrule
\end{tabular}
\end{center}
\vspace{-1.7em}
\end{table}
\section{Related Works}
Ongoing studies have actively explored dual-segment continuum mechanisms and motions \cite{Bajo2012IntegrationAP, Zeng2021MotionCA, Hong2021DevelopmentAV} to optimize surgical continuum manipulators for intricate tasks. They are using rigid tube to connect the continuum manipulator, thus not considering flexible tube. C. Zhang et al. \cite{Zhang2023FlexibleEI} introduces a flexible continuum manipulator based on elastic flexure for endoscopic surgery, demonstrating its capability in ex-vivo environment. This study has unresolved aspects in evaluating the interaction among each segment. Furthermore, while the flexible tube is applied, performance evaluation within curved paths was not conducted in this research.

Previous studies try to compensate hysteresis in continuum manipulators by primarily adopted two approaches. Firstly, compensating for hysteresis through analytical modeling \cite{Do2014HysteresisMA, Do2015NonlinearFM, Kato2016TendondrivenCR, Kim2021ShapeadaptiveHC}, as seen in Lee et al. \cite{Lee2020NonLinearHC} work, where they propose a simplified hysteresis model that defined dead zones and backlash. Analytical hysteresis compensation is effective when properties of the hysteresis model are clear and simple. However, as the complexity of the hysteresis model increases, analytical compensating encounters evident limitations. Secondly, studies have employed learning-based approach \cite{Wang2021ASF}. Kim et al. \cite{Kim2022RecurrentNN} propose a study that effectively combines analytical and deep learning models to predict hysteresis variations based on the shape of Tendon-Sheath Mechanism (TSM). They focus on single pair of tendon-sheath to predict the hysteresis model in the curved pathway.

Hwang et al. \cite{Hwang2020EfficientlyCC} effectively calibrated da Vinci Research Kit (dVRK)\cite{dvrk} based on recurrent neural network. The dVRK handles rigid laparoscopic tools with cable connections less than 0.5m long. However, the proposed continuum manipulator has more complex and amplified hysteresis model due to factors such as coupled motion by dual segments \cite{Zeng2021MotionCA}, hysteresis of TSM \cite{Do2014HysteresisMA}, 2.5m cable-induced amplified hysteresis \cite{cable_hystere}, and cyclic deformation curves of PEEK \cite{Shrestha2016CyclicDA}. These factors make compensating for hysteresis using simple approach challenging \cite{learning_method}. In this paper, we present a hysteresis compensation method for continuum manipulator with long tendon-sheath actuation.

\begin{figure}[t!]
    \centering
    \includegraphics[width=\linewidth]{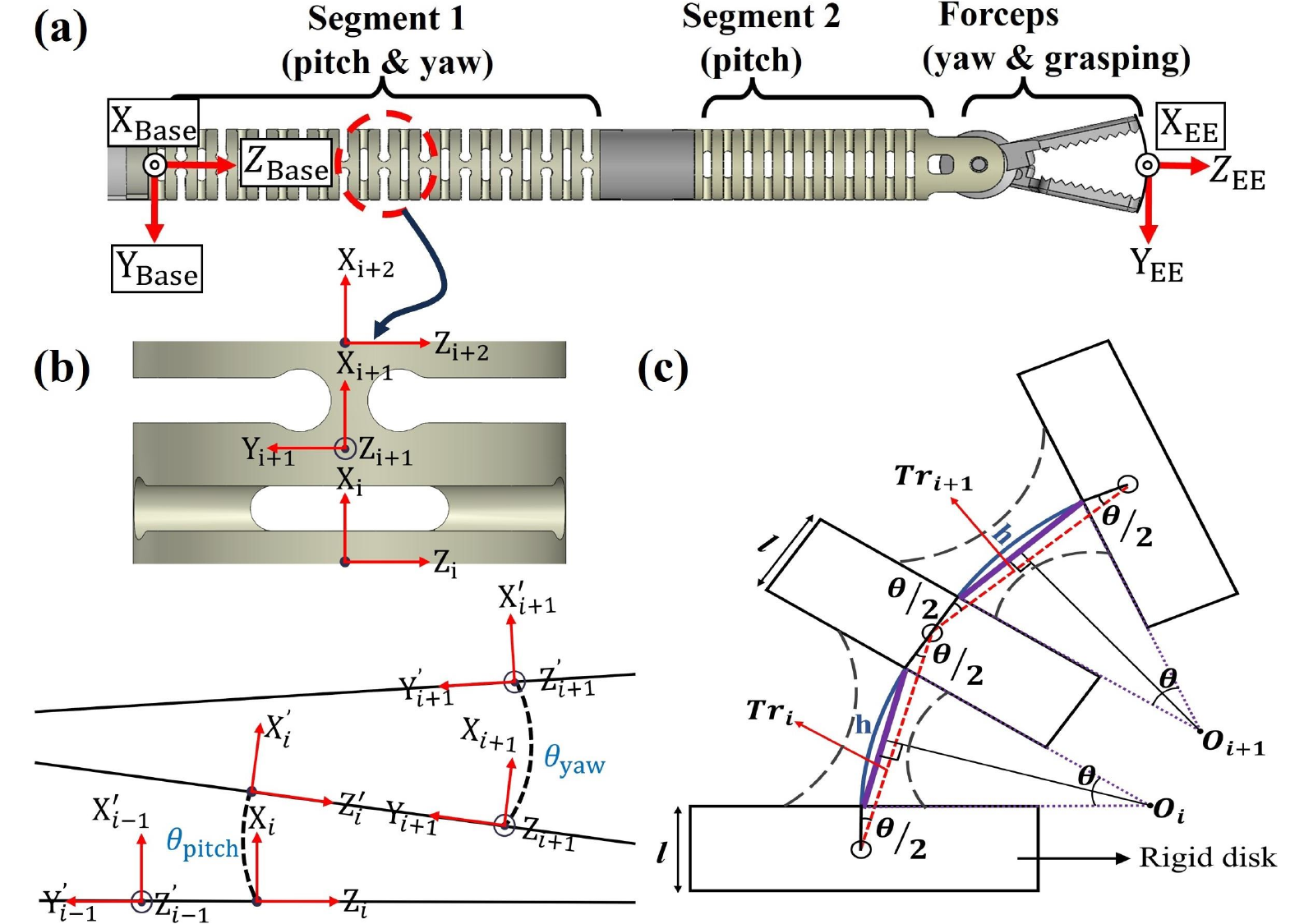}
    \caption{\textbf{Manipulator Components, Coordinate Systems, and Geometric Relationships} (a) The proposed manipulator is constructed by connecting segment 1, segment 2 and forceps. (b) Coordinates attached to a single module (c) Geometric depiction of the flexure hinge module.}
    \vspace{-0.6em}
    \label{Coordinate}
\end{figure}

\begin{table}[t!]
\caption{Kinematics Parameters according to Denavit-Hartenberg Convention for one hinge module of segment 1}
\vspace{-1.6em}
\label{tab:DH_parameter}
\begin{center}
\setlength{\tabcolsep}{5.5pt}
\renewcommand{\arraystretch}{1.3}
\begin{tabular}{c||cccccccc}
\Xhline{3\arrayrulewidth}
\multirow{2}{*}{} & \multicolumn{8}{c}{\textbf{Number of coordinate frame}} \\ 
\cline{2-9} 
                  & \textbf{\textit{1}}     & \textbf{\textit{2}}  & \textbf{\textit{3}}      & \textbf{\textit{4}}  & \textbf{\textit{5}}    & \textbf{\textit{6}}  & \textbf{\textit{7}}  & \textbf{\textit{8}}  \\ \hline
$\alpha_{k-1}$                 & $\pi/2$   & 0  & 0      & 0  & $-\pi/2$  & 0  & 0  & 0  \\ \hline
$a_{k-1}$                 & 0     & 0  & $Tr_i$  & 0  & 0    & 0  & $Tr_{i+1}$  & 0  \\ \hline
$d_k$                 & 0     & 0  & 0      & 0  & 0    & 0  & 0  & 0  \\ \hline
$\theta_k$                 & 0     & $\theta_p/2$  & 0      & $\theta_p/2$  & 0    & $\theta_y/2$  & 0  & $\theta_y/2$  \\ \Xhline{3\arrayrulewidth}
\end{tabular}
\end{center}
\vspace{-0.0em}
\end{table}

\vspace{-0.05em}
\section{Design and Analysis of Continuum Manipulator}
\vspace{-0.05em}
\setlength{\textfloatsep}{2.0pt plus 2.0pt minus 2.0pt}
\begin{figure*}[t!]
    \centering
    \includegraphics[width=0.9\linewidth]{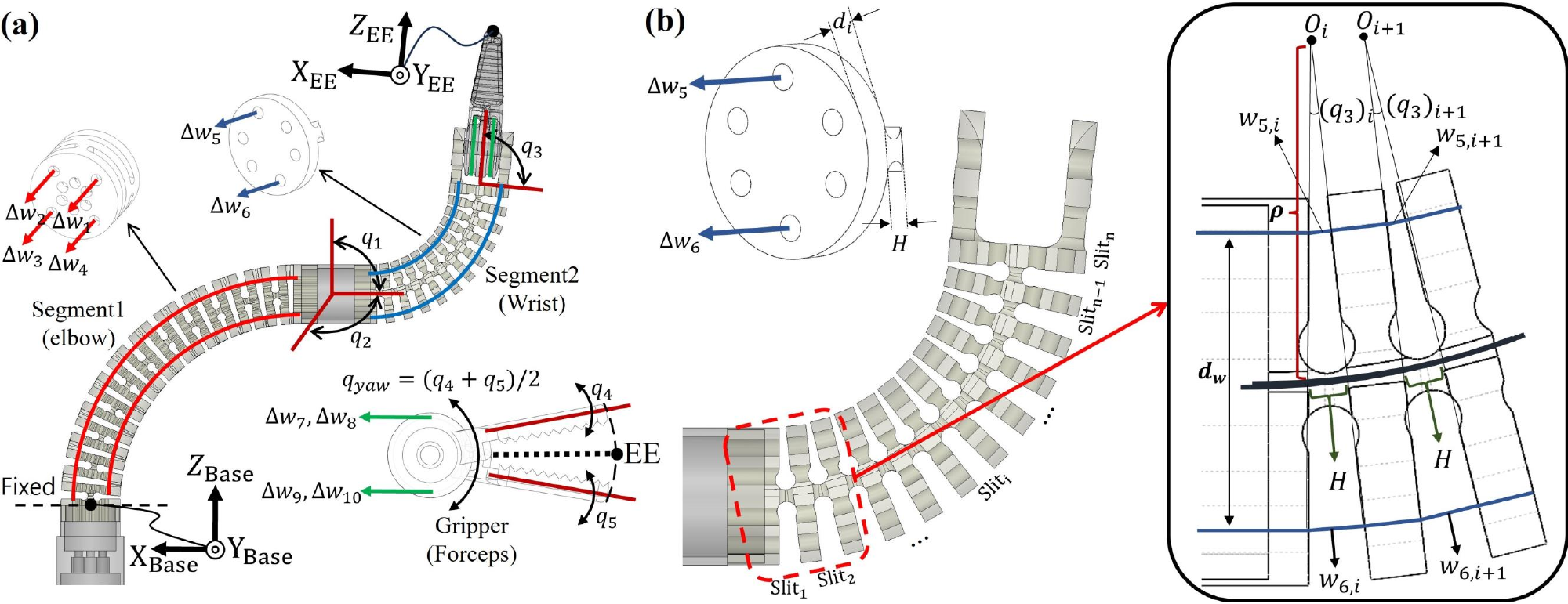}
    \vspace{-0.9em}
    \caption{\textbf{DOFs Configuration and Cable Relationships in The Proposed Continuum Manipulator: (a) Description of the individual DOFs ($q_1$, $q_2$, $q_3$, $q_4$, and $q_5$) for manipulator.} Specifically, segment 1(elbow) has 2 DOFs for pitch and yaw direction bending, driven by cables $w_{1-4}$, denoted as \textbf{$q_1$} and \textbf{$q_2$}, segment 2(wrist) has 1 DOFs for pitch direction bending, driven by cables $w_{5, \: 6}$, represented as \textbf{$q_3$} and the gripper(forceps) is equipped with 2 DOFs, driven by cables $w_{7-10}$, for left forceps angle ($q_4$) and right forceps angle ($q_5$), \textbf{(b) Geometric representation of cable variations $w_i$ during the bending of adjacent hinge} (Magnified view of the red-boxed region showing an ideally bent segment 2).}
    \label{cable_driven}
\vspace{-1.0em}
\end{figure*}
The proposed continuum manipulator is composed of segment 1, segment 2, and gripper arranged from the proximal to the distal ends. The manipulator is equipped with a total of 5 DOFs, comprised of 2 elbow joints, 1 wrist joint, and 2 grippers. We use 10 actuation cables with a length of 2.5m for flexible power transmission connected to the driving unit.
\subsection{Design Considerations}

Flexure hinge-based continuum manipulators offer the advantage of miniaturization due to their simple structure.The specific design parameters are detailed in \Cref{tab:Design_para}. (See \Cref{Design_para}). Finite element analysis was conducted to confirm the suitability of flexure hinge design made of PEEK. The bending angle of a single hinge model is 5.5$^{\circ}$, with a maximum strain of 0.0378. Additionally, when the deformation amplitude ranges from 0.04 to 0.035 at $\mathrm{0.5Hz}$ between 0.04 and 0.035, the number of repetitions ranges from 18454 to 92078 \cite{Shrestha2016CyclicDA}. This generally corresponds to a high cycle life, ranging from \(10^4\)  to \(10^6\). The proposed manipulator consists of a total of 11 hinge modules, maintaining a high cycle life even with a 60.5$^{\circ}$ deflection angle.

\subsection{Kinematics Analysis of Continuum Manipulator}
\label{sec:kinematics}
The composition of the proposed manipulator is described in \Cref{Coordinate}-(a). The bending motion of a flexure-hinge module can be assumed to involve two identically rotating revolute joints and one prismatic joint (See \Cref{Coordinate}-(b)). \Cref{tab:DH_parameter} provides parameters for a single module of segment 1 based on the Modified Denavit-Hartenberg convention. The symbols ${\theta}_{p}$, ${\theta}_{y}$ in \Cref{tab:DH_parameter} denote the bending angles in the pitch, and yaw direction, respectively. Additionally, the bending angle ${Tr}_{i}$ in \Cref{tab:DH_parameter} is calculated using the following relationship (See \Cref{Coordinate}-(c)).
\begin{align}
Tr_i = l\cdot\cos{\frac{\theta_i}{2}} + \frac{2h}{\theta_i}\cdot\sin{\frac{\theta_i}{2}}\quad\quad(i=1,\dots,n) 
\end{align}

The transformation matrix between adjacent equivalent joints can be expressed as follows.
\begin{align}
^{k-1}T_k=
\begin{bmatrix}
   \text{c}\theta_k & -\text{s}\theta_k & 0 & a_{k-1}  \\
   \text{s}\theta_k\text{c}\alpha_{k-1} & \text{c}\theta_k\text{c}\alpha_{k-1} & -\text{s}\alpha_{k-1} & -d_k\text{s}\alpha_{k-1}  \\
   \text{s}\theta_k\text{s}\alpha_{k-1} & \text{c}\theta_k\text{s}\alpha_{k-1} & \text{c}\alpha_{k-1} & d_k\text{c}\alpha_{k-1} \\
   0 & 0 & 0 & 1 \\
   \end{bmatrix}
\end{align}
where c$\theta_k$, s$\theta_k$, c$\alpha_k$, and s$\alpha_k$ are $\cos{\theta_k}$, $\sin{\theta_k}$, $\cos{\alpha_k}$, $\sin{\alpha_k}$, respectively. The transformation matrix for a single module of segment 1 is acquired by multiplying coordinate systems from 1 to 8 in \Cref{tab:DH_parameter}. Similarly, for segment 2's module, it is obtained by multiplying coordinate systems from 2 to 4 in \Cref{tab:DH_parameter}.
\vspace{-0.2em}
\begin{align}
{_{}^{base}}T_{EE}={_{}^0}T_1{_{}^1}T_2{_{}^2}T_3\dots{_{}^{k-1}}T_k\dots{_{}^{n-2}}T_{n-1}{_{}^{n-1}}T_n    
\vspace{-1em}
\end{align}

Concatenating the coordinate systems of all 11 single modules of segment 1 makes the position and orientation of segment 1 (elbow). For segment 2, employ a similar process to interconnect all modules. Ultimately, by multiplying all transformation matrices by considering the rotation of the forceps, the transformation matrix from base to end-effector can be obtained.

We use Newton-Raphson method to obtain the solution for the following inverse kinematics problem:
\vspace{-0.2em}
\begin{gather}
q_{k+1}=q_{k}-J^{-1}(q_{k})(x_{d}-f(q_{k}))\nonumber\\
\underset{\mathbf{q}\in \mathbb{R}^{4}}{\text{minimize}}\:f(\mathbf{q}),\:\:\: \text{where} \: f:\mathbb{R}^{4} \rightarrow \mathbb{R}
\end{gather}
\vspace{-0.2em}
Here, the subscript $k$ denotes the number of iteration, 
$q_{k}$ is the joint configuration, $J_{k}$ is the jacobian matrix, and the objective function $f$ represents forward kinematics function. Specifically, the initial of $q_{k}$ is provided as a command joint configuration. $q_{k}$ consists of a total of 4 values, representing the pitch($q_{1}$) and yaw($q_{2}$) of segment 1, the pitch($q_{3}$) of segment 2, and the yaw direction rotation($q_{yaw} = (q_{4}+q_{5})/2$) of the gripper (See \Cref{cable_driven}-(a)).

\subsection{Cable Drive Equation}
\label{sec:cable_driven}

\Cref{cable_driven}-(b) depicts the geometric relationship showing cable variations during bending at each joint angle. The change in actuating cable length can be expressed as follows.
\begin{align} \label{eq:cable_eq}
\Delta w_i(\theta_i,d_i) = \left(\frac{H\cdot n}{\theta_i}-\frac{d_i}{2}\right)\cdot2\cdot\sin{\frac{\theta_i}{2n}}\quad(i=1,\dots,n)
\end{align}

where $\Delta w_i$ is the variation in cable length, $H$ is the length between notches, $\theta_i$ is the angle at which each hinge bends, and $n$ is the total number of hinges. We control an differential actuation pair of cables to drive a joint based on (5) (See \Cref{cable_driven}). Additionally, we consider decoupling for wrist and gripper control affected by segment bending. Namely, when segment 1 bends in the pitch direction, cables $\mathrm{w}_\mathrm{5}$ and $\mathrm{w}_\mathrm{6}$, controlling segment 2, are adjusted by pulling or releasing them in response to the bending of segment 1.

\section{Hysteresis Modeling and Compensation}

\begin{figure}[t!]
    \centering
    \includegraphics[width=0.85\linewidth]{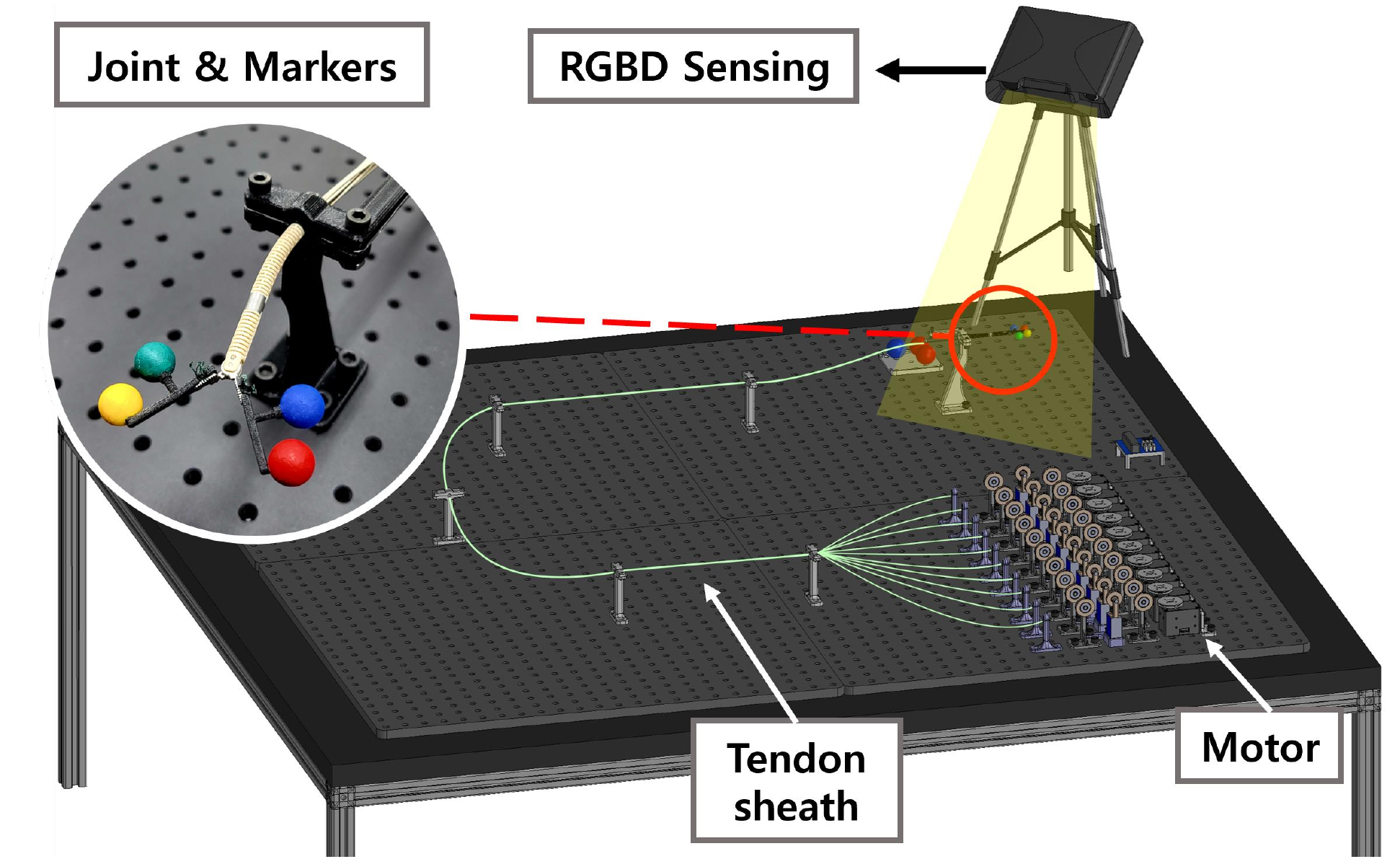}
    \vspace{-0.8em}
    \caption{\textbf{Schematic Illustration of Manipulators with Fiducial Markers and RGBD Sensing:} The proposed manipulator with $\mathrm{4.8 mm}$ in diameter and $\mathrm{2.5 m}$ in length. Seven fiducial markers are attached to capture the physical pose of the manipulator.}
    \label{testbed}
\end{figure}

Due to the hysteresis, there exist a discrepancy between the command joint configurations ($\mathrm{\textbf{q}}_{\mathrm{cmd}}$) and the measured physical joint configurations ($\mathrm{\textbf{q}}_{\mathrm{phy}}$). To model the hysteresis, we measure $\mathrm{\textbf{q}}_{\mathrm{phy}}$ in accordance with $\mathrm{\textbf{q}}_{\mathrm{cmd}}$ using fiducial markers and RGBD sensing (See \Cref{testbed}). In this section, we describe a learning-based method for modeling cabling effects using 2 different approaches and 4 distinct models, and a controller compensating hysteresis by utilizing the trained models.
\subsection{Estimation of Physical Joint Configuration}
\label{sec:estimation_physical}

\begin{figure}[t!]
    \centering
    \includegraphics[width=0.9\linewidth]{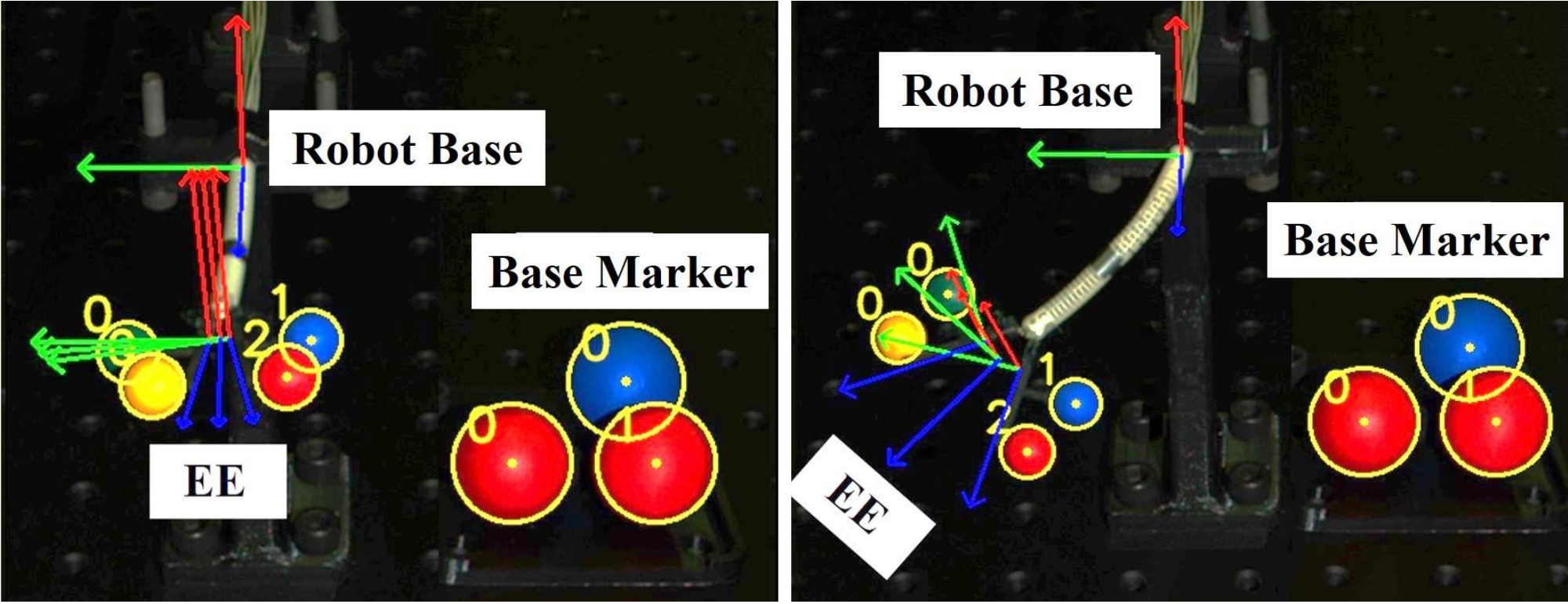}
    \vspace{-0.9em}
    \caption{\textbf{Estimation of Physical Joint Configuration through RGBD Sensing and Fiducial Markers:} Large two red and blue markers are utilized for base pose estimation. The left forceps (${\mathrm{forceps}}_\mathrm{1}$) utilize small red and blue markers, while the right forceps (${\mathrm{forceps}}_\mathrm{2}$) utilize small yellow and green markers for each forceps pose estimation.}
    \label{marker_detect}
\end{figure}

We obtain the physical configuration of the proposed manipulator utilizing colored fiducial markers and RGBD sensing. As illustrated in \Cref{marker_detect}, we employ HSV thresholding to identify each colored marker and obtain its corresponding point clouds. Subsequently, the RANSAC algorithm \cite{ransac} is applied on each point clouds to estimate the center of each marker. We designed the arrangement of markers with different offsets to avoid occlusion in various poses. We assign 2 spheres to each forceps to measure the gripper angle. The transformation matrix from the camera to the base through the positions of the center points of the 3 spheres is obtained as follows.
\vspace{-0.5em}
\setlength{\textfloatsep}{2.0pt plus 2.0pt minus 2.0pt}
\begin{align}
T^{cam}_{base}=\begin{bmatrix}
    R^{cam}_{base} & P^{cam}_{base} \\
    0 & 1
\end{bmatrix}=\begin{bmatrix}
    | & | & | & \\
    {\hat{\text{x}}_B}^{\;\;\text{T}} & {\hat{\text{y}}_B}^{\;\;\text{T}} & {\hat{\text{z}}_B}^{\;\;\text{T}} & p^{cam}_{base} \\
    | & | & | & \\
    0 & 0 & 0 & 1
\end{bmatrix}
\end{align}

where
\begin{align}
\begin{cases}
    \hat{\text{x}}_B = \hat{\text{y}}_B \cross \hat{\text{z}}_B, \: \: \:
    \hat{\text{y}}_B = v^{r_1}_{r_0} / \Vert v^{r_1}_{r_0} \Vert, \: \: \:
    \hat{\text{z}}_B = v^{b_0}_{r_1} / \Vert v^{b_0}_{r_1} \Vert\\
    p^{cam}_{base} = (p_{r_0} + p_{r_1})/2 + L_{base}
\end{cases}
\end{align}

In (7), the subscripts $B$ with $x$, $y$, $z$ denotes the base, while the subscripts $r_0$, $r_1$, and $b_0$ represent the color and number of the sphere and maintain the same notation throughout.$\mathrm{\ }L_{base}$ is the offset between the robot base and the base marker. Next, we determine the orientation and position for the ${\mathrm{forceps}}_1(\text{left})$ and represent the homogeneous transformation matrix as follows.
\vspace{-0.5em}
\begin{align}
T^{cam}_{fcps_1}=\begin{bmatrix}
    R^{cam}_{fcps_1} & P^{cam}_{fcps_1} \\
    0 & 1
\end{bmatrix}=\begin{bmatrix}
    | & | & | & \\
    {\hat{\text{x}}_{f_1}}^{\;\;\text{T}} & {\hat{\text{y}}_{f_1}}^{\;\;\text{T}} & {\hat{\text{z}}_{f_1}}^{\;\;\text{T}} & p^{cam}_{fcps_1} \\
    | & | & | & \\
    0 & 0 & 0 & 1
\end{bmatrix}
\end{align}

where
\begin{align}
\begin{cases}
    \hat{\text{x}}_{f_1} = \frac{v^{b_1}_{r_2}}{\Vert v^{b_1}_{r_2}\Vert} \cross \frac{v^{b_1}_{y_0}}{\Vert v^{b_1}_{y_0} \Vert}, \: \: \:
    \hat{\text{y}}_{f_1} = \hat{\text{x}}_{f_1} \cross \hat{\text{z}}_{f_1}, \: \: \: 
    \hat{\text{z}}_{f_1} = \frac{v^{r_2}_{b_1}}{\Vert v^{r_2}_{b_1} \Vert} \\
    P^{cam}_{fcps_1} = P_{b_1} + L_{fcps_1}
\end{cases}
\end{align}
\vspace{-1.4em}
\begin{align}
R^{cam}_{fcps_1} = R_x(\psi)\cdot R^{cam}_{fcps_1}
\end{align}
\vspace{-1.4em}

In (9) and (10), the subscripts ${f}_\mathrm{1}$ with $x$, $y$, $z$ and ${{fcps}}_1$ denote the left forceps (${\mathrm{forceps}}_1$) and $\psi$ represents the angle resulting from an offset design aimed at reducing marker occlusion. The position and orientation of right forceps (${\mathrm{forceps}}_2$) can be obtained the same as ${\mathrm{forceps}}_\mathrm{1}$. The orientation and position of the end-effector can be obtained as follows.
\vspace{-0.4em}
\begin{gather}
\theta = cos^{-1}(\hat{z}_{f_{1}}\cdot\hat{z}_{f_{2}}) \\
R^{cam}_{EE} = R^{cam}_{fcps_1} \cdot R_x(\theta/2) \\ 
P^{cam}_{EE} = (P^{cam}_{fcps1}+P^{cam}_{fcps2})/2+ d_{fcps1} \cdot (1- cos(\frac{\theta}{2}))\cdot \hat{z}_{f_{1}} \\
T^{base}_{EE} = (T^{cam}_{base})^{-1}\times T^{cam}_{EE}
\end{gather}

where $\mathrm{\theta}$ denote the angle between ${\mathrm{forceps}}_\mathrm{1}$ and ${\mathrm{forceps}}_\mathrm{2}$ and $\mathrm{d}_{{\mathrm{fcps}}_\mathrm{1}}$ denote the distance from the rotation center to the end of ${\mathrm{forceps}}_\mathrm{1}$. From the (14), we calculate $T_{EE}^{base}$ and apply inverse kinematics to obtain the joint configuration. From the $\mathrm{q_{yaw}}$ value in the joint configuration (refer to \Cref{sec:kinematics}), the ${\mathrm{forceps}}_1$ angle ($q_4$) and ${\mathrm{forceps}}_2$ angles ($q_{5}$) can be determined as $q_{yaw}-\theta/2$ and $q_{yaw}+\theta/2$, respectively.

\subsection{Dataset Collection and Hysteresis Observation}

This section details the process of creating a dataset ($\mathzapf{D}$) to capture the hysteresis effect in the manipulator's joint space due to cabling effect has joint-specific attributes. 
\begin{align}
\mathzapf{D} = \left (\left ( \textbf{q}_{\mathrm{cmd}}, \textbf{q}_{\mathrm{phy}} \right )  \right )^{t=1,2,3,...,30707}
\end{align}
\hspace{1em}The $\textbf{q}_\mathrm{cmd}$ was generated through a two-step process. We first randomly selected 1,802 joint configurations within the manipulator's joint limits. These configurations represent a diverse set of poses within reachable workspace. Next, to achieve a uniform distribution with an interval of $3^\circ$ on each of the 5 joints across joint space (resulting in an euclidean distance of $3\sqrt{5}$ between points), interpolation was performed between these randomly chosen configurations, and resulting 30,707 command joint angles. The interpolation method utilized through (16). 
\vspace{-0.3em}
\begin{gather}
\textbf{q}_{\mathrm{p}_{i,j}} = \textbf{q}_{\mathrm{p}_i} + 3j \sqrt{5} \cdot \frac{\textbf{q}_{\mathrm{p}_{i+1}}-\textbf{q}_{\mathrm{p}_i}}{|| \textbf{q}_{\mathrm{p}_{i+1}} - \textbf{q}_{\mathrm{p}_{i}}||} \: \: \: \: (j \leq \lfloor \frac{|| \textbf{q}_{\mathrm{p}_{i+1}} - \textbf{q}_{\mathrm{p}_{i}}||}{3\sqrt{5}} \rfloor)
\end{gather}
where $\textbf{q}_{\mathrm{p}_{i}}$ and $\textbf{q}_{\mathrm{p}_{i+1}}$ are consecutive randomly chosen configuration, $\lfloor \frac{|| \textbf{q}_{\mathrm{p}_{i+1}} - \textbf{q}_{\mathrm{p}_{i}}||}{3\sqrt{5}} \rfloor$ is the number of interpolated points between them, and $|| \cdot ||$ is the $l_2$ norm.
\begin{figure}[t!]
\centering
\includegraphics[width=\linewidth]{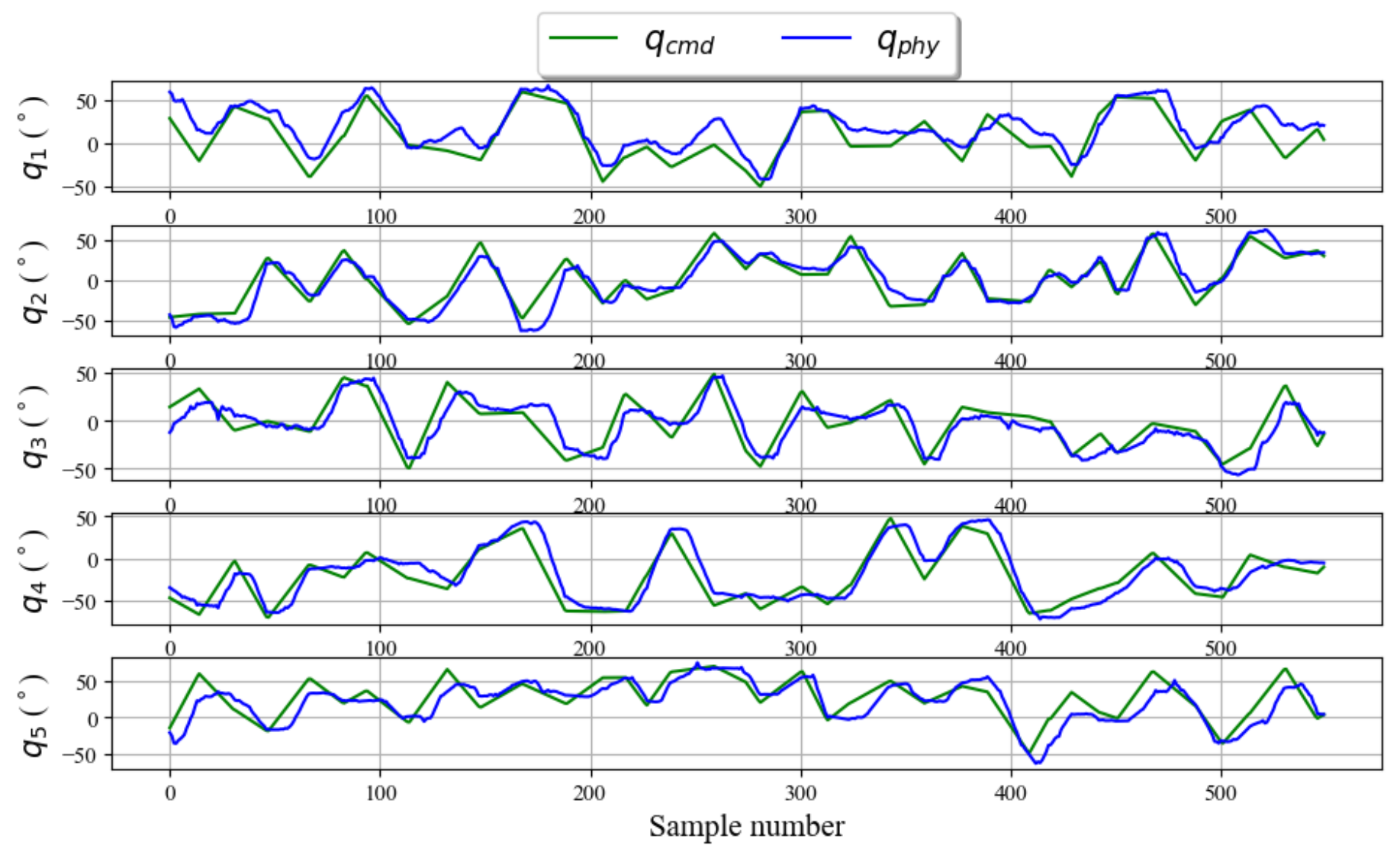} \\
\small{(a) Plotting subset of $\mathbf{q}_{\mathrm{cmd}}$} and $\mathbf{q}_{\mathrm{phy}}$ in joint space\\\vspace{0.3em}
\includegraphics[width=0.9\linewidth]{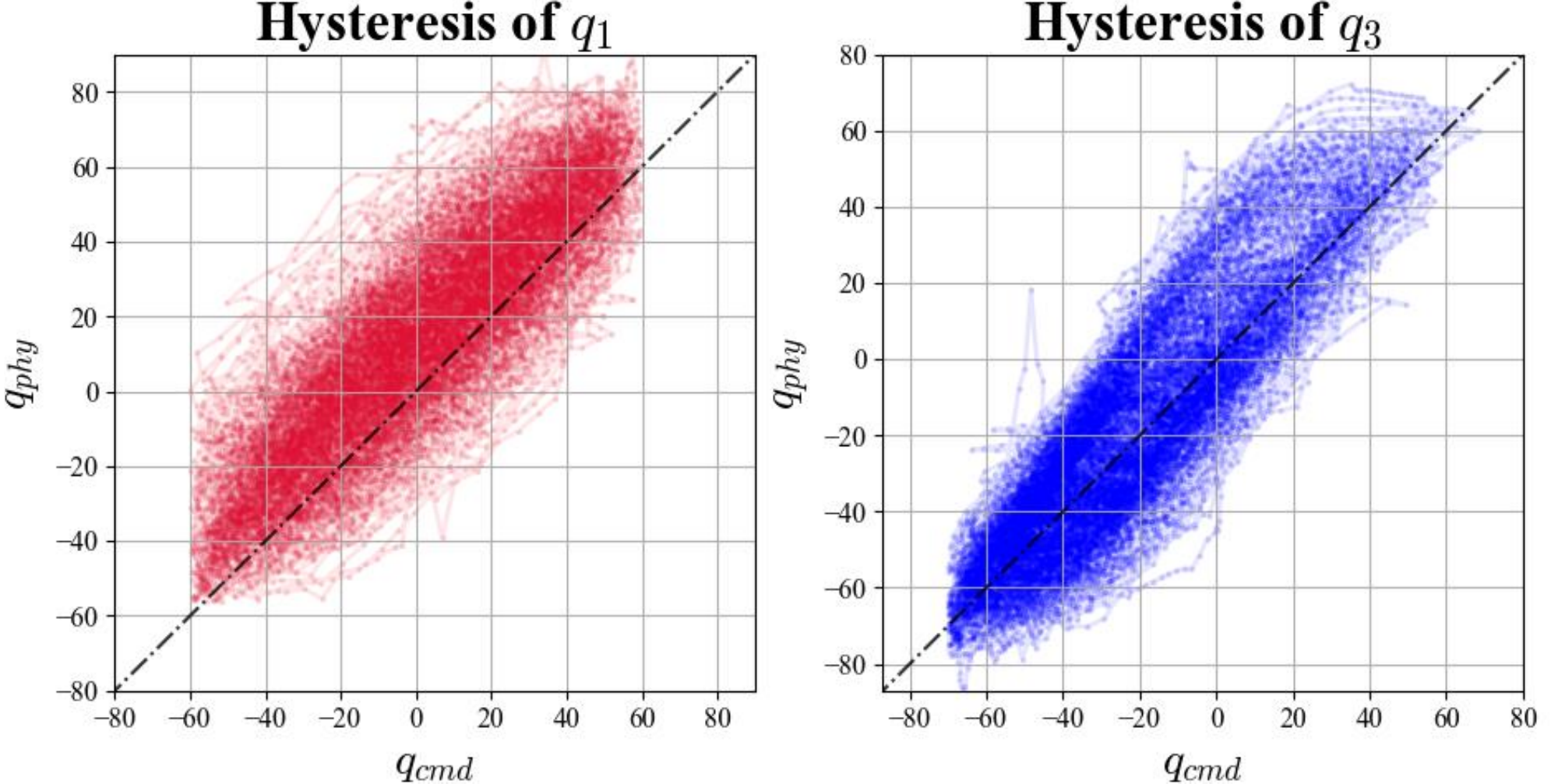} \\
\small{(b) Hysteresis model for $q_1$ and $q_3$ on collected 30,707 dataset}
\vspace{-0.4em}
\caption{\textbf{The Comparison of The Command and The Physical Joint Configurations throughout Collected Dataset $\mathzapf{D}$:} (a) The plot of command and physical joint angles for a subset of collected dataset (e.g., 550 pairs among 30,707 pairs dataset) (b) Observing hysteresis by plotting command joint angle and corresponding physical joint angle on $q_1$ and $q_3$.}
\vspace{-0.1em}
\label{Hysteresis}
\end{figure}
\begin{table}[t!]
\vspace{-0.8em}
\caption{Mean Absolute Error and Standard Deviation between physical and command Joint Configuration on Entire Dataset}
\vspace{-1.8em}
\label{tab:MAE_STD}
\begin{center}
\setlength{\tabcolsep}{10pt}
\renewcommand{\arraystretch}{1.5}
\begin{tabular}{c||ccccc}
\Xhline{3\arrayrulewidth}
\multirow{2}{*}{} & \multicolumn{5}{c}{\textbf{Joint angle}} \\ 
\cline{2-6} 
                  & $q_1$[$^{\circ}$] & $q_2$[$^{\circ}$] & $q_3$[$^{\circ}$] & $q_4$[$^{\circ}$] & $q_5$[$^{\circ}$] \\ \hline
\textbf{MAE}      & 16.31 & 10.09 & 11.21 & 12.02 & 13.84 \\ \hline
\textbf{SD}       & 11.59 & 7.67 & 7.56 & 8.91 & 10.37\\ \Xhline{3\arrayrulewidth}
\end{tabular}
\vspace{-0.9em}
\end{center}
\end{table}
The predefined random command trajectory $\textbf{q}_\mathrm{cmd}$ was used to generate motor commands based on the cable drive equation (5). The robot paused for 0.5 seconds after executing a command joint angle in $\textbf{q}_\mathrm{cmd}$, waiting the vision algorithm (described in \Cref{sec:estimation_physical}) to capture and estimate the corresponding actual physical joint angle ($\textbf{q}_\mathrm{phy}$), resulting total collection time of 256 minutes.

A notable nonlinear difference, commonly referred to as hysteresis, between $\mathbf{q}_{\mathrm{phy}}$ and $\mathbf{q}_{\mathrm{cmd}}$ is evident (See \Cref{Hysteresis}-(a)). The corresponding statistics are summarized in \Cref{tab:MAE_STD}. \Cref{Hysteresis}-(b) illustrates a hysteresis of $q_1$ and $q_3$, where the observed irregularities indicate the challenge of modeling hysteresis in analytic approach.

\subsection{Hysteresis Modeling using Deep Learning}
\label{sec:deeplearning}
In this section, we explores 2 different approaches and 4 distinct architectures to estimate the hysteresis behavior of the proposed manipulator. All approaches and architectures are implemented in PyTorch\cite{pytorch} and utilizing common hyperparameters during training phase, including a 0.001 learning rate, 32 batch size, Mean Square Error (MSE) loss functions, and the utilization of the Adam optimizer.
\begin{figure}[t!]
    \centering
    \includegraphics[width=\linewidth]{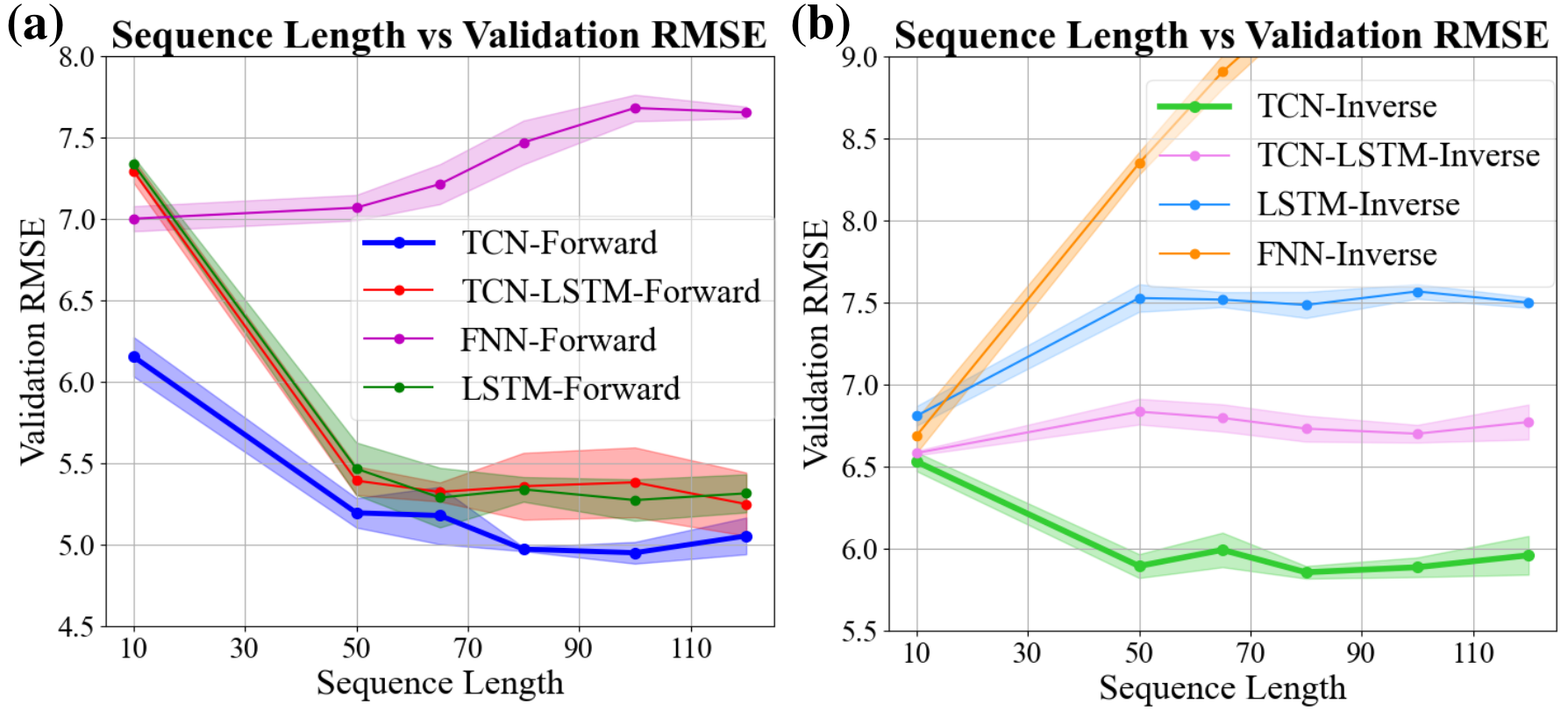}
    \vspace{-1.8em}
    \caption{\textbf{Validation RMSE for Varying Input Sequence Length in Forward and Inverse Model:} The recorded validation RMSE values are presented, with line width indicating SD. (a) Forward model: TCN exhibits optimal performance across overall sequence inputs, whereas TCN-LSTM and LSTM exhibit comparable performance. (b) Inverse model: Most model starts converging at $L = 50$, but FNN model struggles to extract features with an increase of sequence length.}
    \label{forward_loss}
\end{figure}
\noindent \textit{1) Approaches}: we employ the forward approach $(\textit{f}_\mathrm{\theta})$ and the inverse approach $({\textit{f}}_{\theta}^{\: \: -1})$. The forward approach predicts physical joint angle from sequence of command joint angles while the inverse approach predicts command joint angle from sequence of  physical joint angles. In mathematical terms,
\vspace{-0.8em}
\begin{align}
\textbf{Forward: } \mathbf{\hat{q}}^{t}_{\mathrm{phy}} = \textit{f}_{\theta}\left (\textbf{q}^{(t-L, ...,\:t-2,\:t-1, \:t)}_{\mathrm{cmd}} \right ) \\
\textbf{Inverse: } \mathbf{\hat{q}}^{t}_{\mathrm{cmd}} = \textit{f}^{-1}_{\theta}\left (\textbf{q}^{(t-L, ...,\:t-2,\:t-1, \:t)}_{\mathrm{phy}} \right )
\end{align}
where $L$ is the length of input sequence. \newline
\vspace{-0.8em}

\noindent \textit{2) Architectures}: We consider the following candidate architectures for $\textit{f}_\mathrm{\theta}$ and $\textit{f}_\mathrm{\theta}^{\: -1}$.

\textbf{FNN}: A feed-forward neural network (FNN) featuring 3 fully-connected layers with ReLU activation between each layer. The first layer has $5L$ neurons, the second and third layers both have 128 neurons, and the final layer outputs 5 neurons.

\textbf{LSTM}: This architecture employs a single long short-term memory (LSTM) layer \cite{Hochreiter1997LongSM} with $L$ memory cells, each having an input size of 5 and a hidden size of 128. The LSTM layer is followed by 2 fully-connected linear layers with 128 and 64 neurons, respectively, leading to a final output layer with 5 neurons.

\textbf{TCN}: A Temporal Convolutional Network (TCN) \cite{Bai2018AnEE} utilizes series of residual blocks ($R$) as described in (19). Each block comprises with 2 dilated causal convolutions with ReLU activation, and residual connection directly add the input to the block's output. The convolution on each block preceded by left padding to preserve sequence length. We use a kernel size ($k$) of 3 and a dilation base ($d$) of 2. The dilation factor of the convolution on each block exponentially increases the across blocks (e.g., $2^0$, $2^1$, ..., $2^{num \: block-1}$). The number of residual blocks is dynamically determined by the input sequence length ($L$) using the (20) (e.g., $L= 10$ : 2 blocks, $L= 50$ : 4 blocks, and $L = 65, 80, 100, 120$ : 5 blocks). 
\begin{gather}
z_{1,2,3,...,L}^{\: t} =  R_{numblock}\left ( \cdots \left ( R_{2}\left (R_{1} ( \textbf{q}_\mathrm{cmd}^{(t-L\:, ..., \:t)}) \right ) \right ) \cdots \right ) \\
num \: block = \lceil \log_{d} \frac{(L-1)\cdot(d-1)}{2k-2} + 1 \rceil
\end{gather}
The series of residual blocks results in a feature vector ($z_{1,2,3,...,L}^{t}$) as described in (19). We utilized the last feature ($z_{L}^{t}$) for the final output.

\textbf{TCN-LSTM}: A TCN-LSTM model is hybrid model incorporating TCN and LSTM. The feature extracted by TCN ($z_{1,2,3,...,L}^{t}$) serves as input to an LSTM. The setting with TCN and LSTM are the same as described above. This hybrid model outperforms individual TCN and LSTM models, particularly in predicting network traffic \cite{Bi2021AHP} and particulate matter concentration \cite{Ren2023DeepLC}.

To facilitate the training of the models, we partition $\mathzapf{D}$ into an 80\% training set (24,565 pairs) and a 20\% validation set (6,142 pairs). The training process is conducted on an RTX 3070 graphics card, with each model undergoing training for 8,000 epochs. Considering hysteresis dependence on historical states, we vary input sequence lengths ($L$ = 10, 50, 65, 80, 100, 120) to identify optimal lengths. Each of the models is trained 3 times with random weight initialization. We selected the optimal model among the 8,000 epochs and the mean and Standard Deviation (SD) of validation Root Mean Square Error (RMSE) are computed for each model (See \Cref{forward_loss}). 

The TCN model with $L$ = 80 exhibits superior performance for both forward and inverse approaches with the lowest mean and SD of validation RMSE. Despite TCN-LSTM generally performing better, TCN excels in this specific task. Notably, the superior performance of TCN is achieved with significantly fewer trainable parameters compared to LSTM and TCN-LSTM architectures (e.g., at $L$ = 80, LSTM: 77,701, TCN: 800, TCN-LSTM: 78,501). This suggests TCN's effectiveness in capturing temporal features through dilated convolutions, potentially leading to better generalization with fewer parameters.

\begin{figure}[t]
    \centering
    \includegraphics[width=\linewidth]{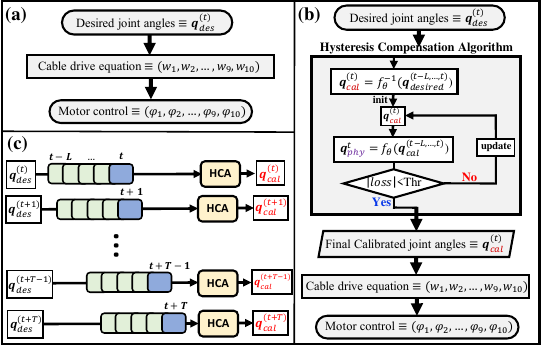}
    \vspace{-1.8em}
    \caption{\textbf{Box Diagram for Uncalibrated and Calibrated Controller} (a) Uncalibrated controller (b) Calibrated controller utilizing the hysteresis compensation algorithm (c) Conceptual image for utilizing calibrated controller on inputted $\textbf{q}_\mathrm{desired}^{(t)},\dots,\textbf{q}_\mathrm{desired}^{(t+T)}$. The HCA represents the proposed hysteresis compensation algorithm.}
    \vspace{0.5em}
    \label{fig:controller}
\end{figure}

\begin{algorithm}[t!]
\caption{Hysteresis Compensation Algorithm}
\label{algo1}
\begin{algorithmic}[1]
    \REQUIRE Desired joint angle $\textbf{q}^{(t)}_\mathrm{desired}$, TCN-forward $f_{\theta}$, TCN-inverse $f^{\: -1}_{\theta}$, number of iterations \textit{M}, learning rate $\alpha$, time sequence length \textit{L}, loss threshold \textit{Thr}, joint size \textit{Q}
    \STATE ${\textbf{q}^{(t)}_\mathrm{calibrated}}$ $\leftarrow$ $f^{-1}_{\theta}({\textbf{q}^{(t-L, \dots, t-1, t)}_\mathrm{desired}})$
    \STATE \textbf{for} iteration $\in$ $\{1, \dots, M\}$
    \STATE \quad loss $\leftarrow$ $f_{\theta}({\textbf{q}^{(t-L, \dots, \: t-1, \: t)}_\mathrm{calibrated})}\:-\:$ ${\textbf{q}^{(t)}_\mathrm{desired}}$ 
    \STATE \quad \textbf{for} idx $\in$ $\{1, \dots, Q\}$
    \STATE \quad \quad \textbf{do if} $\left\vert \mathrm{loss[idx]} \right\vert$ $>$ $Thr$
    \STATE \quad \quad \quad \textbf{then} ${\textbf{q}^{(t)}_\mathrm{calibrated}}$ [idx] $\leftarrow$ $\textbf{q}^{(t)}_\mathrm{calibrated}$ [idx] $- \: \alpha\cdot$ loss
    \RETURN ${\textbf{q}^{(t)}_\mathrm{calibrated}}$
\end{algorithmic}
\end{algorithm}

\subsection{Design of Hysteresis Compensation Algorithm}
\begin{figure*}[t!]
    \centering
    \includegraphics[width=0.9\linewidth]{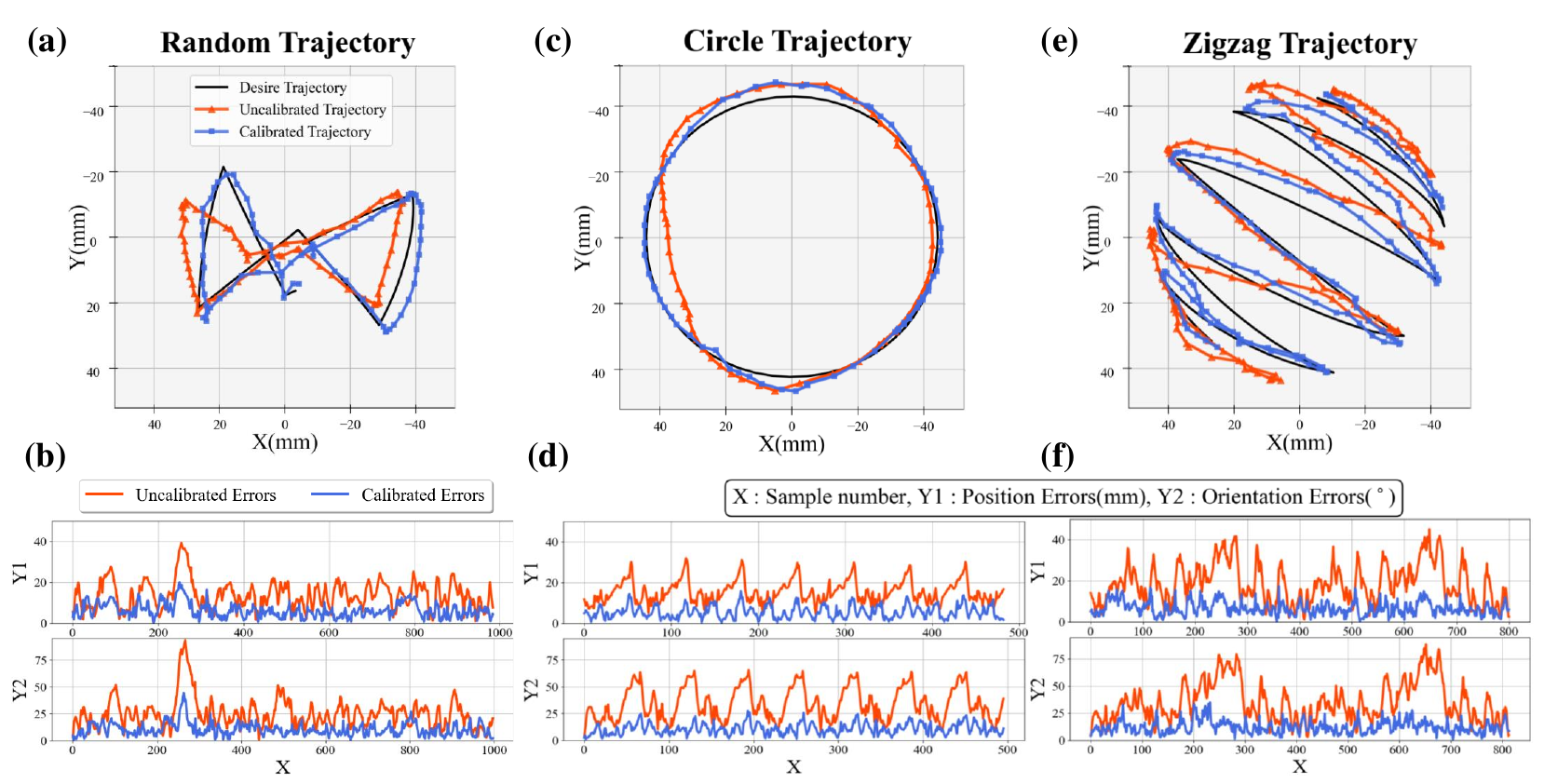}
    \vspace{-1em}
    \caption{\textbf{The Performance Comparison between Uncalibrated and Calibrated Controllers in terms of Position and Orientation Errors during Trajectory Tracking Tests: } Subfigures (a)-(b) show performance and errors on a random trajectory, while (c)-(d) demonstrate performance and errors on a circular trajectory and lastly (e)-(f) show performance and errors on a zigzag trajectory. Before calibration, errors in random trajectories show unpredictable deviations overall. In the circular trajectory, the error indicated a repetitive pattern of increasing values over a particular interval. In the zigzag trajectory, the error revealed the most inconsistent large deviation. However, after calibration, all trajectories showed a significantly reduced and uniform deviation. For visibility, only specific parts of the trajectories are shown in (a), (c), and (e).}
    \vspace{-0.0em}
    \label{Trajectory_val}
\end{figure*}

\begin{table*}
\caption{Position and Orientation Errors and SD for Uncalibrated and Calibrated Control across Multiple Trajectories in Task Space}
\vspace{-1.8em}
\renewcommand{\arraystretch}{1.0}
\label{tab:Validation_table}
\begin{center}
\begin{tabular}{c|cc|cc|cc|cc} 
\hline
\multirow{2}{*}{\begin{tabular}[c]{@{}c@{}}MAE on~\\unseen trajectory\end{tabular}} & \multicolumn{2}{c|}{Random Trajectory}                      & \multicolumn{2}{c|}{Circle Trajectory}  & \multicolumn{2}{c|}{Zigzag Trajectory}  & \multicolumn{2}{c}{Average Value}        \\ 
\cline{2-9}
                                                                                    & uncalibrated                       & calibrated             & uncalibrated   & calibrated             & uncalibrated   & calibrated             & uncalibrated   & calibrated              \\ 
\hline
X Error [$\mathrm{mm}$]                                                             & 7.44                               & \textbf{3.27}          & 6.66           & \textbf{2.31}          & 8.84           & \textbf{2.66}          & 7.65           & \textbf{2.75}           \\
Y Error [$\mathrm{mm}$]                                                             & 7.39                               & \textbf{3.60}          & 7.04           & \textbf{2.48}          & 7.34           & \textbf{3.40}          & 7.26           & \textbf{3.16}           \\
Z Error [$\mathrm{mm}$]                                                             & 5.54                               & \textbf{2.44}          & 9.34           & \textbf{3.62}          & 11.37          & \textbf{3.63}          & 8.75           & \textbf{3.23}           \\ 
\hline
Total Position Error [$\mathrm{mm}$]                                                & \multicolumn{1}{l}{11.86$\pm$6.8}  & \textbf{5.47$\pm$3.5}  & 13.46$\pm$5.4  & \textbf{4.96$\pm$3.1}  & 16.16$\pm$9.4  & \textbf{5.64$\pm$3.3}  & 13.70$\pm$7.2  & \textbf{5.29$\pm$3.3}   \\ 
\hline
Orientation Error [$^{\circ}$]                                                      & \multicolumn{1}{l}{24.74$\pm$13.9} & \textbf{10.41$\pm$5.9} & 33.88$\pm$15.2 & \textbf{10.65$\pm$5.1} & 34.89$\pm$18.1 & \textbf{12.58$\pm$5.9} & 31.17$\pm$15.7 & \textbf{11.21$\pm$5.6}  \\
\hline
\end{tabular}
\vspace{-2.0em}
\end{center}
\end{table*}
This section introduces the proposed hysteresis compensation algorithm, a pivotal component of the resulting calibrated controller (See \Cref{fig:controller}-(b)). The primary objective of the compensation algorithm is returning the calibrated command joint angles ($\textbf{q}_\mathrm{calibrated}^{(t)}$) that can achieve the desired joint angles ($\textbf{q}_\mathrm{desired}^{(t)}$). This is accomplished by estimating the manipulator's hysteresis behavior using the trained TCN-forward ($f_{\theta}$) and inverse ($f_{\theta}^{\:-1}$) models with $L = 80$ from \Cref{sec:deeplearning}. The hysteresis compensation algorithm (refer to \Cref{algo1}) operates in two phases:

\textbf{1) Initializing $\textbf{q}_\mathrm{calibrated}^{(t)}$ with TCN-inverse }: The inverse model, $f_{\theta}^{-1}$, estimates command joint configurations from the physical joint configurations. Consequently, feeding $\textbf{q}_\mathrm{desired}^{(t-L,\dots, t-1, t)}$ into $f_{\theta}^{-1}$ should ideally return the calibrated command configuration that would result in the $\textbf{q}_\mathrm{desired}^{(t)}$.
 
\textbf{2) Refining $\textbf{q}_\mathrm{calibrated}^{(t)}$ using TCN-forward }: Given the higher RMSE of the inverse model compared to the forward approach, we refine the initial estimate $\textbf{q}_\mathrm{calibrated}^{(t)}$ using $f_{\theta}$. The forward model predicts physical joint angles based on command joint angles. Feeding the current estimate of $\textbf{q}_\mathrm{calibrated}^{(t)}$ into $f_{\theta}$ provides an estimate of the resulting physical joint configurations. We define the loss (21) based on the difference between the desired and predicted physical joint configurations, computed on the current estimate of $\textbf{q}_\mathrm{calibrated}^{(t)}$.
\vspace{-1em}
\begin{gather}
    \mathrm{loss} = f_{\theta}({\textbf{q}^{(t-L, \dots, \: t-1, \: t)}_\mathrm{calibrated})}\:-\: {\textbf{q}^{(t)}_\mathrm{desired}}
\end{gather}

A positive loss indicates overshooting, while a negative loss indicates undershooting. $\textbf{q}_\mathrm{calibrated}^{(t)}$ is iteratively updated by subtracting $\alpha \cdot \mathrm{loss}$ until the absolute value of individual joint losses fall below a predefined loss threshold ($Thr$), or until a predefined number of iterations ($M$) is reached.

Unlike the uncalibrated controller that directly calculates motor commands (in \Cref{sec:cable_driven}) from desired joint angles (See \Cref{fig:controller}-(a)), the calibrated controller (See \Cref{fig:controller}-(b)) incorporates the hysteresis compensation algorithm. This refines the $\textbf{q}_\mathrm{desired}^{(t)}$ to account for the manipulator's hysteresis, resulting in $\textbf{q}_\mathrm{calibrated}^{(t)}$.

\Cref{fig:controller}-(c) illustrates control method utilizing the proposed calibrated controller. When the user commands $\textbf{q}_\mathrm{desired}^{(t)}$, the controller automatically attaches the history of commanded desired joint angles ($\textbf{q}_\mathrm{desired}^{(t-L,\dots, t-1)}$) to the current desired angle. If the history is less than $L-1$ steps, zero-padding is applied to the left side of the sequence for consistency. This history, along with the current desired angle, is fed into the hysteresis compensation algorithm to obtain the $\textbf{q}_\mathrm{calibrated}^{(t)}$.
\vspace{-0.2em}

\section{Results and Validation}

\subsection{Trajectory Tracking Test}
To compare the performance of the proposed calibrated controller against uncalibrated controller, a tracking experiment was conducted on 3 unseen trajectories (random, circular, and zigzag) in task space. In configuring the controller for the proposed manipulator, parameters are set as $Q$ = 5, $M$ = 50, $\alpha$ = 0.001, and $Thr$ = 4. Errors and SDs are calculated by performing the trajectory 5 times. 

The tracking performance across different target trajectories is visualized in \Cref{Trajectory_val}, with detailed mean absolute error and SD values for position and orientation presented in \Cref{tab:Validation_table}. The uncalibrated controller resulted in a substantial average position error of $\mathrm{13.70 \: mm}$ for the 3 trajectories. In comparison, the calibrated controller significantly reduced the error to $\mathrm{5.29 \: mm}$. The mean and SD of the position error were approximately 61\% and 54\% reduced by leveraging calibrated controller.

\begin{figure}[!t]
\centering
\includegraphics[width=0.93\linewidth]{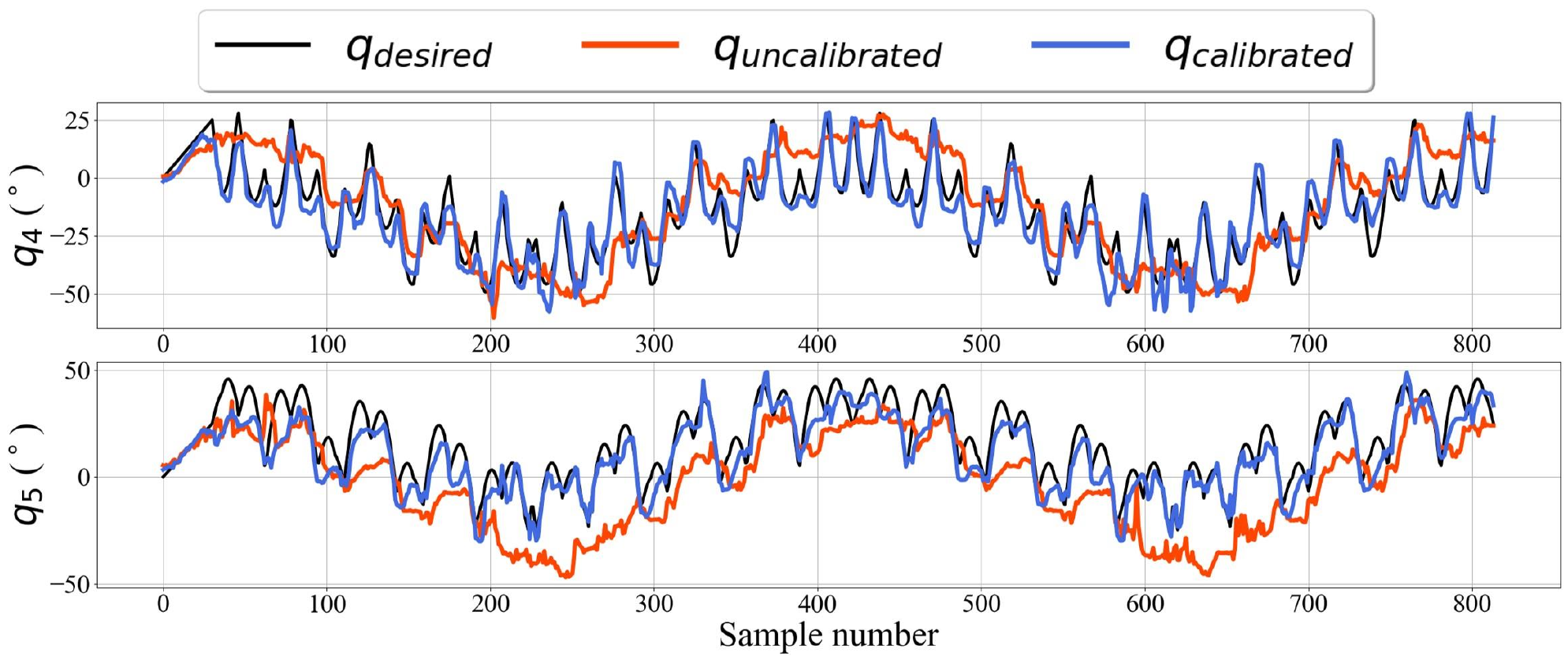} \\
\small{(a) The tracking performance comparison between calibrated and uncalibrated controller for ${\mathrm{q}}_\mathrm{4}$ and ${\mathrm{q}}_\mathrm{5}$ in joint space}
\includegraphics[width=0.93\linewidth]{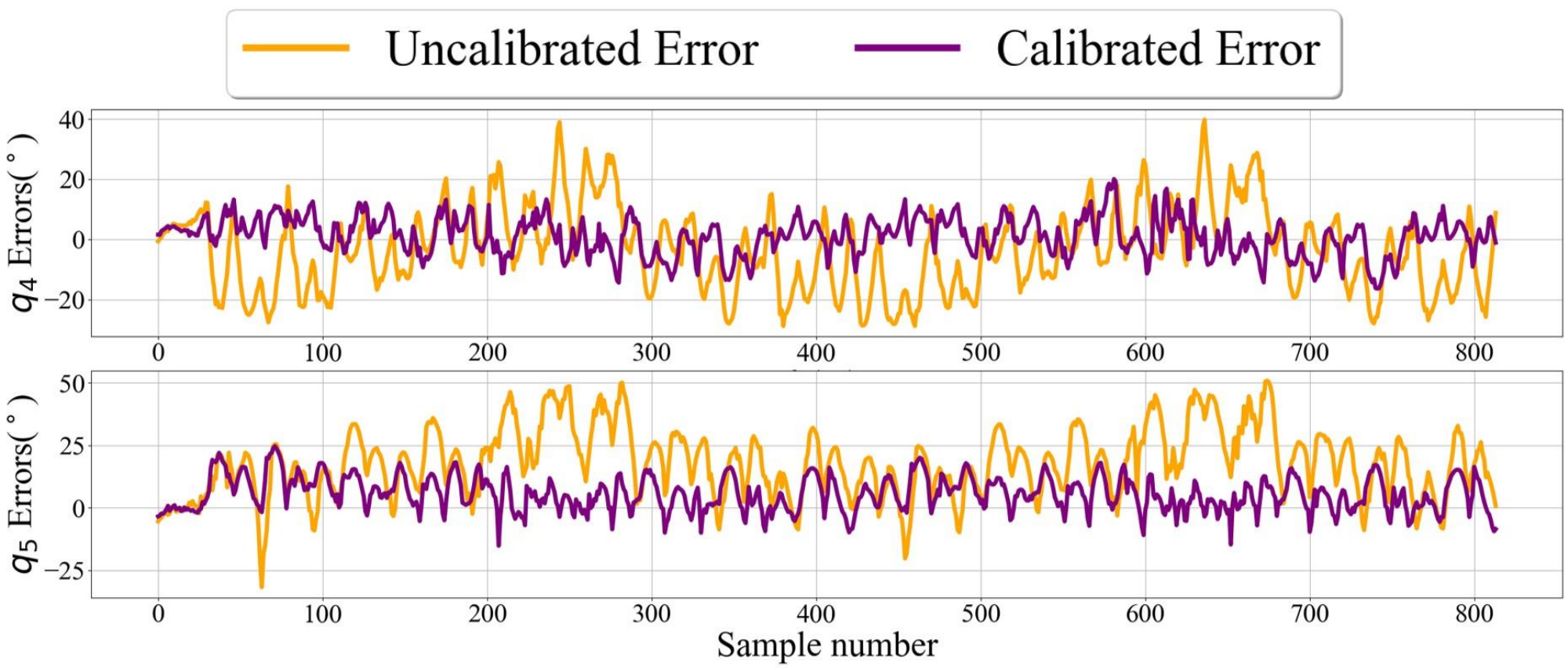}\\
\small{(b) The comparison of tracking error for ${\mathrm{q}}_\mathrm{4}$ and ${\mathrm{q}}_\mathrm{5}$ in joint space}
\vspace{-0.8em}
\caption{\textbf{Enhanced Results in Forceps Angle ($\mathrm{q}_4$, $\mathrm{q}_5$) at The Joint Space (Zigzag trajectory)}}
\label{forcep_err}
\vspace{-0.9em}
\end{figure}

Particularly noteworthy was the significant improvement in the forceps movement among all joint angles during the zigzag trajectory. As illustrated in \Cref{forcep_err}, the calibrated controller effectively compensates for hysteresis during intricate movements of the forceps. Specific statistics of tracking errors are documented in \Cref{Joint_Angle_Error}. 

According to \cite{Kim2020EffectOB}, hysteresis exceeding 10° can significantly impact surgical task completion time. The proposed compensation controller is expected to reduce the hysteresis to less than 10° on average, potentially improving surgical performance.
\vspace{-0.2em}
\section{Conclusions}
\vspace{-0.1em}
In this paper, we proposed a 5-DOF flexible continuum manipulator for endoscopic surgery. However, the designed manipulator accompanies significant hysteresis. To address this, we proposed a deep learning-based control algorithm.
The method utilizes RGBD sensing to track 7 fiducial markers and gather a dataset for model learning. With a controller based on the best model (e.g., TCN at $L = 80$), we perform an unseen trajectory tracking test. The results demonstrated a significant reduction in mean errors in both task space and joint space compared to uncalibrated controller that suggests potentially shorter surgery completion times and higher precision control during surgical tasks.

In future work, we will apply the proposed calibration method to high precision surgical tasks such as suturing and anastomosis. The limitation of this study includes the required time to collect the dataset, approximately 256 minutes. Future research aims to efficient calibration method by faster surgical tool pose estimation using stereo cameras. In practical surgical applications, prolonged use may result in cable slackening \cite{cable_slack}, which is not captured in the training data. To address this issue, we are considering updating the model by fine-tuning it from a pre-trained model with a small number of additional datasets collected during instances of cable slackening.
\begin{table}[t!]
\caption{Joint Angle Errors and Standard Deviations under Uncalibrated and Calibrated Control at Zigzag Trajectory}
\vspace{-0.8em}
\label{Joint_Angle_Error}
\begin{tabularx}{\linewidth}{@{}c *{6}{>{\centering\arraybackslash}X}@{}}
\toprule
\multicolumn{2}{l}{\multirow{2}{*}{}} & \multicolumn{5}{c}{\textbf{Joint Angle Error}} \\
\cmidrule(l){2-7} 
\multicolumn{2}{l}{} & $q_1$[$^{\circ}$] & $q_2$[$^{\circ}$] & $q_3$[$^{\circ}$] & $q_4$[$^{\circ}$] & $q_5$[$^{\circ}$] \\
\midrule
\multirow{2}{*}{\textbf{Uncalibrated control}} & \textbf{MSE} & 15.60 & 13.73 & 12.97 & 11.37 & 18.97 \\
 & \textbf{SD} & 11.42 & 10.54 & 8.50 & 8.19 & 12.15 \\
\midrule
\multirow{2}{*}{\textbf{Calibrated control}} & \textbf{MSE} & \textbf{4.67} & \textbf{6.37} & \textbf{7.52} & \textbf{5.21} & \textbf{6.99} \\
 & \textbf{SD} & \textbf{3.96} & \textbf{4.80} & \textbf{5.82} & \textbf{3.80} & \textbf{5.39} \\
\bottomrule
\end{tabularx}
\vspace{-0.1em}
\end{table}

\addtolength{\textheight}{-12cm}   

\bibliographystyle{IEEEtran}
\bibliography{ral_ref}

\end{document}